\PassOptionsToPackage{dvipsnames,table}{xcolor}
\documentclass[sigconf]{acmart}

\copyrightyear{2023}
\acmYear{2023}
\setcopyright{acmlicensed}
\acmConference[MM '23] {Proceedings of the 31st ACM International Conference on Multimedia}{October 29--November 3, 2023}{Ottawa, ON, Canada.}
\acmBooktitle{Proceedings of the 31st ACM International Conference on Multimedia (MM '23), October 29--November 3, 2023, Ottawa, ON, Canada}
\acmPrice{15.00}
\acmISBN{979-8-4007-0108-5/23/10}
\acmDOI{10.1145/3581783.3613848}
\usepackage{multirow}
\usepackage{amsmath}
\usepackage[dvipsnames]{xcolor}

\usepackage{amssymb}
\usepackage{makecell}
\usepackage{utfsym}
\usepackage{pifont}
\usepackage{microtype}
\usepackage{graphicx}
\usepackage{subfigure}
\usepackage{booktabs} 
\usepackage{enumitem}

\begin{document}

\title{Retrieval-based Knowledge Augmented Vision Language Pre-training}

\author{Jiahua Rao}
\authornote{Both authors contributed equally to this research.}
\affiliation{
  \institution{School of Computer Science and Engineering, Sun Yat-sen University}
  \city{Guangzhou}
  \country{China}
}
\email{raojh6@mail2.sysu.edu.cn}

\author{Zifei Shan}
\authornotemark[1]
\affiliation{%
  \institution{WeChat, Tencent}
  \city{Shenzhen}
  \country{China}
  }
\email{zifeishan@tencent.com}

\author{Longpo Liu}
\affiliation{%
  \institution{WeChat, Tencent}
  \city{Shenzhen}
  \country{China}
}
\email{longpoliu@tencent.com}

\author{Yao Zhou}
\affiliation{
 \institution{WeChat, Tencent}
  \city{Shenzhen}
  \country{China}
 }
\email{yaoozhou@tencent.com}

\author{Yuedong Yang}
\authornote{Corresponding author.}
\affiliation{%
  \institution{School of Computer Science and Engineering, Sun Yat-sen University}
  \state{Guangzhou}
  \country{China}
}
\email{yangyd25@mail.sysu.edu.cn}







\begin{abstract}
    With the recent progress in large-scale vision and language representation learning, Vision Language Pre-training (VLP) models have achieved promising improvements on various multi-modal downstream tasks. Albeit powerful, these models have not fully leveraged world knowledge to their advantage. 
    A key challenge of knowledge-augmented VLP is the lack of clear connections between knowledge and multi-modal data. Moreover, not all knowledge present in images/texts is useful, therefore prior approaches often struggle to effectively integrate knowledge, visual, and textual information.
    In this study, we propose \textbf{RE}trieval-based knowledge \textbf{A}ugmented \textbf{V}ision \textbf{L}anguage (\textbf{REAVL}), a novel knowledge-augmented pre-training framework to address the above issues.
    For the first time, we introduce a knowledge-aware self-supervised learning scheme that efficiently establishes the correspondence between knowledge and multi-modal data, and identifies informative knowledge to improve the modeling of alignment and interactions between visual and textual modalities.
    By adaptively integrating informative knowledge with visual and textual information, REAVL achieves new state-of-the-art performance uniformly on knowledge-based vision-language understanding and multi-modal entity linking tasks, as well as competitive results on general vision-language tasks while only using 0.2\% pre-training data of the best models. 
    Our model shows strong sample efficiency and effective knowledge utilization.
\end{abstract}







\maketitle

\section{Introduction}

Recent Vision-Language Pre-training (VLP) models, such as ALBEF \cite{li2021align}, BLIP \cite{li2022blip} and SimVLM \cite{wang2021simvlm}, learn multi-modal representations from large-scale image-text pairs via well-designed pre-training objectives. However, despite the strong performance of these models, they generally cannot well incorporate external world knowledge in pre-training, requiring increasingly larger networks or data to cover more facts with huge computational costs.
Meanwhile, large knowledge graphs (KGs), such as Wikidata \cite{vrandevcic2014wikidata} and ConceptNet \cite{speer2017conceptnet}, can provide structured world knowledge to multi-modal data by representing entity descriptions and their relationships, which is implicit in vision/text but comprises complementary  information \cite{wang2021kepler,yasunaga2022dragon}. Therefore, leveraging structured knowledge and multi-step reasoning in KGs can really complement multi-modal learning.

However, building connections between knowledge and multi-modal data remains a stiff challenge. 
While previous studies such as KEPLER \cite{wang2021kepler}, and DRAGON \cite{yasunaga2022dragon} have incorporated knowledge into pre-training models explicitly, most of them are restricted by the knowledge behind the textual modality, neglecting a large amount of knowledge in other modalities like images. They often require entity-linking tools to build connections between knowledge and text.
More importantly, these models have not fully leveraged world knowledge to their advantage because there is no ground truth indicating which knowledge is helpful for multi-modal representation.
MuRAG and REVEAL \cite{chen2022murag,hu2023reveal} proposed to assess the usefulness of retrieved knowledge from images for language generation modeling, but they still struggle with redundant and useless knowledge.
Consequently, existing works may limit their potential to incorporate informative knowledge into multi-modal data.

To this end, we propose a \textbf{RE}trieval-based knowledge \textbf{A}ugmented \textbf{V}ision-\textbf{L}anguage Pre-training (REAVL) model, which retrieves world knowledge from KGs and leverage them to improve the multi-modal representations. REAVL has two core components: a knowledge retriever that retrieves the informative knowledge given multi-modal data, and a knowledge-augmented model that fuses multi-modal data and knowledge. Concretely, we take multi-modal data and a KG as the input data, and first retrieve the related knowledge from the KG via the knowledge retriever. Then we use a knowledge-augmented model to encode these inputs into fused representations, where the multi-modal features are fused with knowledge features through the cross-attention layer in an adaptive way.

To efficiently incorporate informative knowledge into multi-modal representations, we pretrain our model by introducing a knowledge-aware self-supervised learning scheme including four different types of modules: masked language modeling (K-MLM), masked vision modeling (K-MVM), image-text contrastive learning (K-ITC), and KG link prediction (LinkPred). 
Specifically, K-MLM and K-MVM are the knowledge-aware masked data modeling objectives, where the original signal is reconstructed by using its masked modal inputs and the corresponding retrieved knowledge. The masked learning signals help to reward helpful knowledge and penalize uninformative ones during knowledge retrieval. And K-ITC focuses on aligning images and text in multi-modal data.
Furthermore, LinkPred makes the model use the KG structure jointly with the corresponding multi-modal data to reason about missing links in KG.
We conduct ablation studies and show the effectiveness and complementarity of individual self-supervised tasks.

We benchmark REAVL on multiple vision-language benchmarks, including general, knowledge-based, and multimodal entity linking (MEL) tasks. 
REAVL achieves state-of-the-art performance on knowledge-based and MEL tasks.
Notably, on knowledge-based VQA benchmarks, REAVL achieves new state-of-the-art results while utilizing fewer parameters and less knowledge than previous works. This demonstrates the superiority of our approach for retrieving and augmenting knowledge in multi-modal representations.
On general tasks, REAVL also scores the best within models trained with a similar amount of data, competitive to models trained on billions of images while using only 0.2\% of their data, showing the strong sample efficiency of our methodology.

To summarize, we make the following contributions:
\begin{itemize}[leftmargin=*]
\setlength\itemsep{0mm}
    \item We proposed a novel retrieval-augmented pre-training framework that explicitly leverages large-scale knowledge graphs to assist vision-language pre-training.
    \item We are the first to introduce a knowledge-aware self-supervised learning scheme that efficiently identifies informative knowledge to improve the multi-modal modeling.
    \item We achieved new state-of-the-art performance uniformly on knowledge-based vision-language understanding and multi-modal entity linking tasks, as well as competitive results on general tasks while only using 0.2\% pre-training data of the best models.
\end{itemize}

\section{Related Work}
\subsection{Vision-Language Pre-training Model}
Vision-Language Pre-training has been developing rapidly in recent years. Most existing work focuses on modeling the interactions between images and text with transformer-based encoders\cite{li2021align,li2022blip, zheng2022cross}. And multiple well-designed cross-modal pre-training objectives have been proposed, for example, image-text matching, masked data modeling, and contrastive learning. They have achieved their impressive performance by scaling up the dataset with noisy image-text pairs collected from the web. For example, BLIP \cite{li2022blip} effectively utilizes the noisy web data by bootstrapping the captions, where a captioner generates synthetic captions and a filter removes the noisy ones. 
Recent methods such as SimVLM \cite{wang2021simvlm}, OFA \cite{wang2022ofa}, CoCa \cite{yu2022coca}, and PaLi \cite{chen2022pali} unified the multi-modal pre-training model and self-supervised tasks to obtain state-of-the-art performance on various downstream tasks. These methods or frameworks benefit from their learning ability on the increasing web-collected pre-training data \cite{changpinyo2021cc12m,jia2021scaling}, which also creates the challenge of huge computational costs, and thus complicates the optimization procedure. 

Our work, by contrast, focuses on incorporating world knowledge into multi-modal representations, which have been demonstrated helpful and necessary for various downstream tasks. By introducing knowledge into multi-modal learning, our model can achieve better performance with low-resource pre-training data.

\subsection{Knowledge Augmented Language Model}
Knowledge integration is active research for improving language models. One line of research aims to add entity features into language models (LMs) \cite{zhang2019ernie,wang2021kepler,chen2021kb}.
For example, ERNIE \cite{zhang2019ernie} enhanced the text representations with their corresponding named entity mentions at the text token level while KB-VLP \cite{chen2021kb} augmented the knowledge from image-text pairs based on entity-linking tools and object detection models.
A recent method DRAGON \cite{yasunaga2022dragon} bidirectionally fuses text and KG via a deep cross-modal model and learns joint reasoning over text and KG. 
These methods usually require entity linking tools to link text and KG entities before training, and the retrieved knowledge is not updated during training.

Another line of work is retrieval-augmented LMs \cite{guu2020retrieval,lewis2020retrieval,chen2022align}, which retrieve relevant text from a corpus and integrate it into LMs as additional knowledge. For example, REALM \cite{guu2020retrieval} integrates Wikipedia passages as a knowledge base to benefit downstream knowledge-intensive tasks.
However, current retrieval-augmented methods are often restricted to using text-only knowledge, neglecting an amount of knowledge in other modalities like images, which contain much information not covered by text. Compared to them, MuRAG \cite{chen2022murag} is the first retrieval model that is capable of retrieving knowledge presented in multiple modalities while REVEAL \cite{hu2023reveal} construct a large-scale memory by encoding various sources of multimodal world knowledge, including Wikipedia passage and web images with alt-text captions.

To improve on the above works, we propose to effectively retrieve informative knowledge from multiple modalities and to incorporate knowledge with multi-modal representations by knowledge-aware self-supervision learning. The fundamental rationale of our method is that it improves the retrieval steps through well-designed self-supervised learning tasks, which improves the correspondence between knowledge and multi-modal data. Another distinction is that the existing retrieval-based models typically focus on adding entity- or triplet-level knowledge rather than the subgraphs around the retrieved entity to enable richer knowledge fusion.

\subsection{Knowledge Graph representation learning}
In recent years knowledge embeddings have been extensively studied by predicting missing links in graphs. Link prediction is a fundamental task in KGs \cite{kazemi2018simple, zheng2021pharmkg}, and various works study methods to learn KG entity and relation embeddings for link prediction, such as TransE, DistMult, and RotatE \cite{bordes2013translating,yang2014embedding,sun2018rotate,mai2021communicative}. 
Several works additionally use textual data or pre-training LMs to help learn KG embeddings and link prediction. While these works focus on the KG-side representations, we extend the scope and use the KG-side objective (link prediction) jointly with multimodal-side objectives to train a knowledge-augmented vision-language model.

\begin{figure*}[ht]
\begin{center}
\centerline{\includegraphics[scale=0.53]{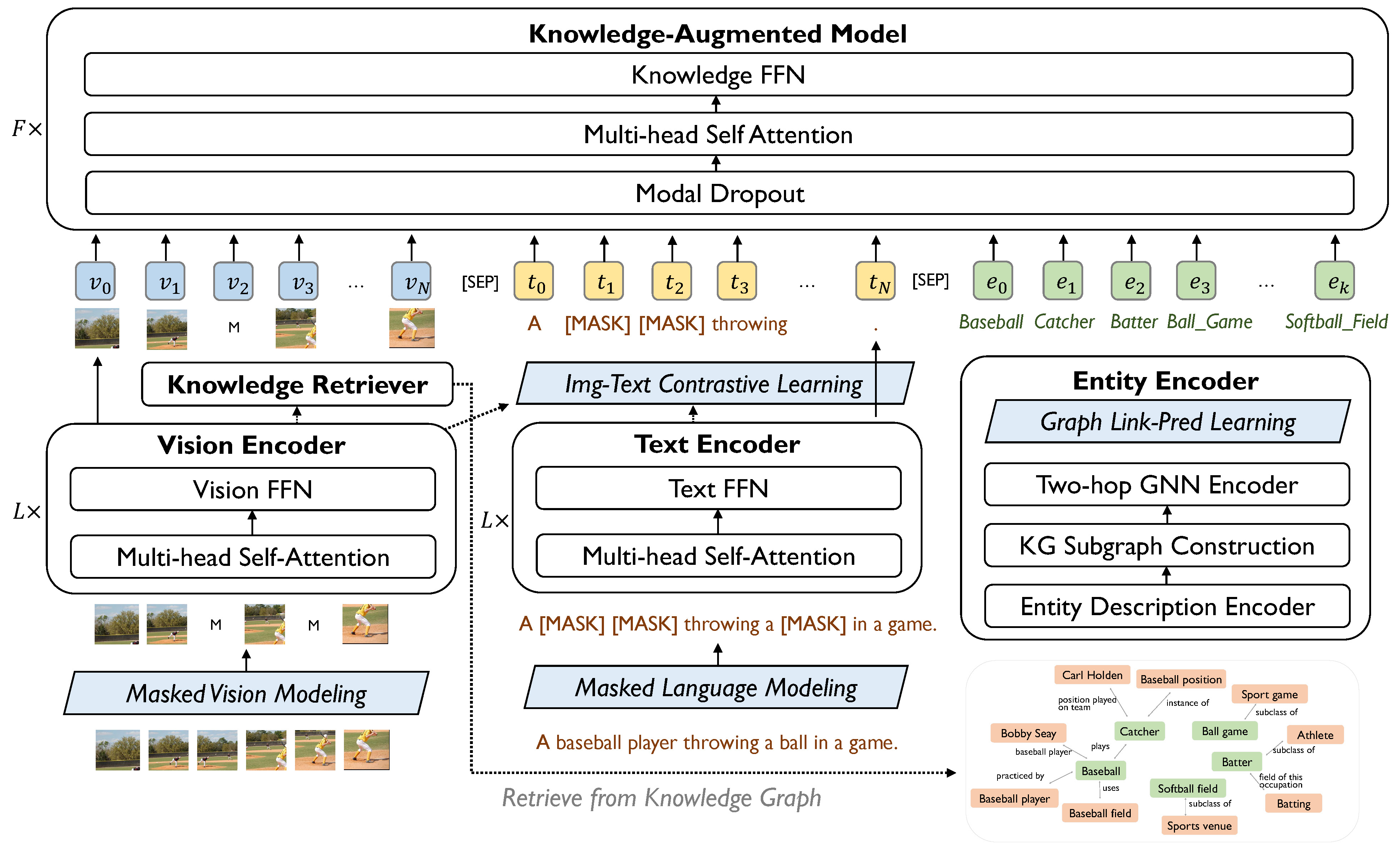}}
\caption{Illustration of the REAVL model. Given multi-modal data and a large knowledge graph (KG), we utilize a Knowledge Retriever to retrieve knowledge from KG from multi-modal data and then incorporate the retrieved knowledge in multi-modal representation learning with a Knowledge-Augmented Model. We also unify four types of knowledge-aware self-supervised tasks (i.e. Masked Language Modeling) to encourage multi-modal data and KG to mutually inform each other.}
\label{icml-framework}
\end{center}
\end{figure*}

\section{Methods}

In this section, we first briefly introduce the model architecture of our approach in Section \ref{sec:arch} and Figure \ref{icml-framework}. Then we describe the proposed Knowledge Retrieval, followed by the GNN Encoder for the retrieved knowledge (Section \ref{sec:retrieve}-\ref{sec:gnn}). In Section \ref{sec:kae}, we show how to effectively incorporate multi-modal data with knowledge by the Knowledge-Augmented Model. Lastly, we describe the pre-training objectives with corresponding self-supervised tasks (Section \ref{sec:obj}).

\subsection{Model Architecture \label{sec:arch}}

At the core of our method is the idea of incorporating multi-modal representations with KGs. 
To achieve this, We defined the image-text pairs as $ \mathcal{D} = \{(I, T) |  I \in \mathcal{I}, T \in \mathcal{T}\}$, and the KG is a multi-relational graph $\mathcal{G} = \{ ( \mathcal{V}, \mathcal{E}, \mathcal{R}  ) \} $ where $\mathcal{V}$ is the set of entity nodes, $\mathcal{E}$ is the set of edges (triplets) and $\mathcal{R}$ is the set of relation types in the KG.

We first employ a vision encoder $f_v$ and a text encoder $f_t$ to produce the representations of the input image $I$ and text $T$. For the vision encoder, we use a 12-layer visual transformer ViT-B/16 \cite{dosovitskiy2020vit}. An input image $I$ is first divided into $16 \times 16$ patches and encoded into a sequence of patch embeddings $\upsilon = \{\upsilon_1, ..., \upsilon_N \} $ where $N$ is the number of image patches. 
The text encoder, a 12-layer transformer, is initialized using the pre-trained BERT$_{base}$ \cite{devlin2018bert} model and transforms an input text $T$ into a sequence of embeddings $t = \{ t_{cls}, t_1, ..., t_{N_t} \} $, where $N_t$ is the number of text tokens and $t_{cls}$ is the encoding of the start token of a sentence.

The patch embeddings $\upsilon$ from image $I$ are then used for knowledge retrieval, which is to find the nearest entities behind the image. The set of retrieved entities is defined as $M \subset \mathcal{V}$. We then add their one-hop neighbor nodes to construct the entity subgraph $G \subset \mathcal{G} $ and feed them into the GNN Encoder. Therefore, we could obtain the retrieved entity embedding $e = \{ e_1, ..., e_k \}$ after aggregating their neighbor information.

Finally, the image and text embedding as well as entity embedding are further processed by the knowledge-augmented model. The knowledge-augmented model is initialized by the first and last layer of the BERT, which enhances multi-modal representation by interacting with other modalities through cross-attention blocks.

\subsection{Knowledge Retrieval \label{sec:retrieve}}

Instead of directly linking entity mentions in the input text to entity nodes in KGs \cite{yasunaga2022dragon}, we followed the strategy of REALM \cite{guu2020retrieval} by explicitly asking the model to decide what knowledge to retrieve and incorporate.

In the retrieval step, the image patches with fine-grained descriptions are helpful for understanding image content and retrieving possibly helpful entities $M$ from the KG $\mathcal{G}$.
Therefore, with the $N$ patches of the input image $I$, the vision encoder $f_v$ is applied to map each patch to a dense representation $\upsilon = \{\upsilon_0, \upsilon_1, ..., \upsilon_N \} $. For knowledge entries in KGs $m \in \mathcal{V}$ ($\mathcal{V}$ is the set of entity nodes), we apply the backbone encoder $f_e$ to encode the entity $m$ with its description. Formally, we define the relevance score between the image patch $v_i$ and the entity $m$ as:

\begin{equation}
    sim(v_i, m_j) = f_v(v_i)^T f_e(m_j) 
\end{equation}

In total, we retrieve the top $ N \times K$ knowledge entries relevant to image $I$. We keep top-k knowledge entries ranked by similarity scores as the set of retrieved entities $M = \{m_1, m_2, ..., m_k\}$.

In practice, we use CLIP model \cite{radford2021learning} to encode all of the entity descriptions as the candidate memory and index them using FAISS \cite{johnson2019billion} to find the top-K nearest neighbors for each image patch.

For each resulting set of retrieved entities, we further add their one-hop neighbor nodes from KG to construct the entity subgraph $G$. The subgraph contains richer entities and relations, enabling our model to obtain more information about world knowledge. 

\subsection{GNN Encoder \label{sec:gnn}} 

After extracting the 2-hop entity subgraph around the image-text pair, we applied a GNN encoder to model the representation of informative entities by aggregating their neighbor information and strengthening the relational information, inspired by \cite{yasunaga2022dragon,ijcai2022p544}.
To represent the graph, we first obtain initial node embeddings $\{e_1^{(0)}, ..., e_k^{(0)} \}$ for the retrieved entities using the entity encoder.
The initial embedding of the relations in the subgraph is also initialized by their description embeddings.

Then, in each layer of the GNN, the current representation of the node embeddings $\{e_1^{(l-1)}, ..., e_k^{(l-1)} \}$ is fed into the layer to perform a round of information propagation between nodes in the graph and yield pre-fused node embeddings for each entity:
\begin{equation}
    \begin{split}
    \{e_1^{(l)}, ..., e_k^{(l)} \} = {\rm GNN}(\{e_1^{(l-1)}, ..., e_k^{(l-1)} \}) 
    \end{split}
\end{equation}
where GNN corresponds to a variant of graph attention networks and $l$ corresponds to the layer of GNN. 
The GNN computes node representations for each node via message passing between neighbors on the graph:

\begin{equation}
    e_i^{(l)} = f_{n} \left(\sum_{e_j \in \mathcal{N}_{e_i} \cup {e_i}} \alpha_{ij} \cdot f_{A_{ij}} \right) + e_i^{(l-1)}
\end{equation}
where $f_{n}$ is a linear transformation, $\mathcal{N}_{e_i}$ represents the neighborhood of an entity node $e_i$, and $f_{A_{ij}}$ denotes the message passing function from its neighbors $e_j$ to $e_i$. The $\alpha_{ij}$ is an attention weight that scales the message function $f_{A_{ij}}$, which could be calculated as:
 \begin{equation}
     \beta_{ij} = \frac{f_q (e_i^{(l-1)})^T f_k(e_j^{(l-1)}, r_{ij})}{\sqrt{D}}
 \end{equation}
 
 \begin{equation}
     \alpha_{ij} = \frac{exp (\beta_{ij})}{\sum_{e_k \in \mathcal{N}_{e_i} \cup {e_i}} exp(\beta_{ik})}
 \end{equation}
 where $f_q$ and $f_k$ are linear transformations and $r_{ij}$ is a relation embedding for the relation connecting $e_i$ and $e_j$.

The message passing function $f_{A_{ij}}$ could be computed in the following manner:

\begin{equation}
    f_{A_{ij}} =  f_m (e_j^{(l-1)}, r_{ij}) 
\end{equation}
where $f_m$ is linear transformations and $r_{ij}$ is the same as above.

\subsection{Knowledge Augmented Model \label{sec:kae}}

After using a visual transformer layer, a text transformer layer, and a GNN layer for vision, text, and entities respectively, we use a knowledge-augmented model to let the two modalities fuse the knowledge information through a 2-layer cross-modalities transformer \cite{khare2021mmbert,zheng2020predicting}. 

In principle, we concatenate the pre-fused embeddings including visual embeddings $\upsilon$, text embeddings $t$, and entity embeddings $e$ as the input elements in the BERT model. 
The input sequence always starts with a special classification element ([CLS]), then goes on with visual patch embeddings, then follows up with text token elements, and ends with the retrieved entities. A special separation element ([SEP]) is inserted between the visual and text elements, and between the text and entity elements. Therefore, four types of input elements are involved, namely, visual, textual, entity, and special elements for disambiguating different input formats. Finally, the multi-modal features are fused with knowledge features through the cross-attention block at each layer. 

Let $x = \{x_1, ..., x_p \}$ be the input elements, which are embedding features of the visual, textual, and entity elements. They are processed by a multi-layer bidirectional Transformer, where the embedding features of each element are transformed layer-by-layer in the fashion of aggregating features from the other elements with adaptive attention weights. The features of the $(l + 1)$-th layer, $x^{(l+1)}$, is computed by:

\begin{equation}
    \label{eq:multi-head}
    \widehat{h}_i^{(l+1)} = \sum_{m=1}^{M} W_m^{(l+1)}\left( \sum_{j=1}^{N} A_{i,j}^m \cdot V_m^{(l+1)} x_j^{l}
    \right)
\end{equation}

\begin{equation}
    h_i^{(l+1)} = {\rm LayerNorm}\left(x_i^{l} + \widehat{h}_i^{(l+1)}\right)
\end{equation}

\begin{equation}
   \widehat{x}_i^{(l+1)} = W_2^{(l+1)} \cdot {\rm GELU}\left(W_1^{(l+1)} h_i^{(l+1)}\right) 
\end{equation}

\begin{equation}
   x_i^{(l+1)} = {\rm LayerNorm}\left( h_i^{(l+1)} + \widehat{x}_i^{(l+1)}\right)
\end{equation}
where $m$ in Eq.\ref{eq:multi-head} indexes over the attention heads, and $A_{i,j}^m = {\rm exp}[ (Q_m^{(l+1)} x_i^{l})^T (K_m^{(l+1)} x_j^{l})]  $ denotes the attention weights between elements $i$ and $j$ in the $m$-th head, and $W_m^{(l+1)}$, $Q_m^{(l+1)}$, $K_m^{(l+1)}$ and $W_1^{(l+1)}$, $W_2^{(l+1)}$ are learnable weights for $m$-th attention head and $(l+1)$-th layer.

\subsection{Pre-training objective \label{sec:obj}} 
To ensure that the multi-modal and KG mutually inform each other, we unify four knowledge-aware self-supervised tasks: masked language modeling, masked vision modeling, KG link prediction, and image-text contrastive learning.

\subsubsection{\textbf{Knowledge-aware Masked Language Modeling (K-MLM)}} MLM is a common pre-training task used for language models.  Following the strategy of REALM \cite{guu2020retrieval}, we use Span Masking Modeling to focus on the span $s$ that may require world knowledge to predict the masked tokens.
Let $\widehat{T}$ denote a masked text, and $p^{msk}(I, \widehat{T}, E)$ denote the model’s predicted probability for a masked token. We adopt the strategy of \cite{li2022blip}, which minimizes a cross-entropy loss:

\begin{equation}
    \mathcal{L}_{MLM} = \mathbb{E}_{(I, \widehat{T}, E)} [H(y^{msk}, p^{msk}(I, \widehat{T}, E))]
\end{equation}
where $y^{msk}$ is a one-hot vocabulary distribution where the ground-truth token has a probability of 1.

\subsubsection{\textbf{Knowledge-aware Masked Vision Modeling (K-MVM)}} For image masking, unlike MAE \cite{he2022masked}, we aim to reconstruct the invisible patches features with visible image, text, and entity features to facilitate multi-modal information and knowledge fusion.  Concretely, we propose to apply a reconstruction loss to facilitate the aggregation of multi-modal and knowledge features. Formally, the reconstruction loss is defined as:

\begin{equation}
    \mathcal{L}_{MVM} = \mathbb{E}_{(\widehat{I}, T, E)} [H(y^{msk}, p^{msk}(\widehat{I}, T, E))]
\end{equation}
where $\widehat{I}$  denotes the masked patches and $p^{msk}(\widehat{I}, T, E))$ denotes the predicted probability for masked image patch features.

\subsubsection{\textbf{Knowledge-aware Image-Text Contrastive Learning (K-ITC)}} Image-Text Contrastive Learning aims to learn better unimodal representations before fusion. Inspired by ALBEF \cite{li2021align}, for each image and text, we calculate the softmax-normalized image-to-text and text-to-image similarity as:

\begin{equation}
    s^{(i2t)}(I, T) = \frac{exp(s\langle I, T_m \rangle / \tau) }{\sum_{k \ne i}^K exp(s\langle I, T_m \rangle)}
\end{equation}

\begin{equation}
    s^{(t2i)}(I, T) = \frac{exp(s\langle I_m, T \rangle / \tau) }{\sum_{k \ne i}^K exp(s\langle I_m, T \rangle)}
\end{equation}

\begin{equation}
    \mathcal{L}_{ITC} = \frac{1}{2} \mathbb{E} \left( [H(y^{i2t}, s^{i2t}(I, T)) + H(y^{t2i}, s^{t2i}(I, T))
    \right)
\end{equation}

\subsubsection{\textbf{KG Link Prediction (LinkPred)}} While the MLM and MVM task predicts for the multi-modal and knowledge side, link prediction holds out some edges and predicts them for the input KG. As our approach takes a joint image-text and KG entities pair as input, we expect that link prediction can encourage the model to learn to use the KG structure jointly with the textual context and visual regions to reason about missing links in the KG.    

Concretely, we hold out a subset of edge triplets from the input KG $S = \{ (h,r,t) \} \subset E$. Then we maps each entity node ($h$ or $t$) and relation ($r$) in the KG to a vector, $h$, $t$, $r$, and defines a scoring function $\phi_r(h, t)$ to model positive/negative triplets. Herein, we consider a KG triplet scoring function $\phi_r(h, t)$ as DistMult \cite{yang2014embedding}: $\phi_r(h, t) = <h, r, t>$. $ <\cdot, \cdot, \cdot>$ denotes the trilinear dot product. A higher $\phi$ indicates a
higher chance of $(h, r, t)$ being a positive triplet (edge) instead of negative (no edge). For training, we optimize the objective:

\begin{equation}
\begin{split}
    \mathcal{L}_{LinkPred} = \sum_{(h,r,t) \in S} 
        - {\rm log} \sigma(\phi_r(h, t) + \gamma)\\
        + \frac{1}{n} \sum_{(h^{'}, r, t^{'})} {\rm log \sigma(\phi_r(h^{'}, t^{'} + \gamma)}
\end{split}
\end{equation}
where $(h^{'}, r, t^{'})$ are $n$ negative samples corresponding to the positive triplet $(h, r, t)$, $\gamma$ is the margin, and $\sigma$ is the sigmoid function.

\begin{table*}[t]
\caption{Results for vision-language pre-training methods on popular VL benchmarks. We report accuracy for OK-VQA and SNLI-VE and vqa-score for AOK-VQA and VQA-v2. The best and second-best results are marked {\bf number} and \underline{number}, respectively. The gray \textcolor{lightgray}{number} indicates that the model is trained with a significantly larger number of data than our models. }
\label{res-table}
\vskip 0.05in
\begin{center}
\begin{normalsize}
\begin{tabular}{c|c|lc|cc}
    \toprule
    Task & Dataset & Method & Knowledge Sources & \multicolumn{2}{c}{Results} \\
     & & &  & dev & test \\
     
    \midrule
     \multirow{15}{*}{Knowledge-based Task} 
     & \multirow{8}{*}{OK-VQA} & \textsc{Supervised} & & & \\
     & & KAT & Wikidata + Frozen GPT-3 & \multicolumn{2}{c}{53.1} \\
     & & REVIVE & Wikidata + Frozen GPT-3 & \multicolumn{2}{c}{\underline{56.6}} \\
     \cmidrule{3-6}
     & & \textsc{Pre-training} &  \\
     & & ALBEF & \# image 12M  & \multicolumn{2}{c}{54.7} \\
     & & BLIP & \# image 129M   & \multicolumn{2}{c}{55.4} \\
     & & REVEAL$_{Base}$  & CC12M + Wikidata + WIT + VQA-v2 & \multicolumn{2}{c}{55.2} \\
     & & REAVL (ours) & \# image 4M + Wikidata & \multicolumn{2}{c}{\textbf{57.7}} \\
    \cmidrule{2-6}

     & \multirow{6}{*}{AOK-VQA} & \textsc{Pre-training} & & &  \\
     & & ALBEF & \# images 12M  & \multicolumn{1}{c}{54.5} & - \\
     & & BLIP & \# images 129M  & \multicolumn{1}{c}{\underline{56.2}} & 50.1 \\
     & & REVEAL$_{Base}$  & CC12M + Wikidata + WIT + VQA-v2 & - & 50.4 \\
     & & REAVL (ours) & \# images 4M + Wikidata & \multicolumn{1}{c}{\textbf{58.4}}  & {\textbf{52.7}} \\
    \midrule

     \multirow{22}{*}{General Task} 
     & \multirow{13}{*}{VQA-v2} & \textsc{Base Data-Size} &  & & \\
     & & VL-BERT & \# images 3.3M & 71.16 & - \\
     & & UNITER  &  \# images 4M  & 72.70 & 72.91 \\
     & & OSCAR   &  \# images 4M  & 73.16 & 73.44 \\
     & & UNIMO   &  \# images 4M  & \underline{75.06} & \underline{75.27} \\
     & & ALBEF   &  \# images 4M  & 74.54 & 74.70 \\
     & & REAVL (ours) & \# images 4M + Wikidata &  {\textbf{77.62}} & {\textbf{77.79}} \\
     \cmidrule{3-6}
     & & \textsc{Large Data-Size} &  &  \\
     & & ALBEF   & \# images 12M  & 75.84 & 76.04 \\
     & & BLIP & \# images 14M  & 77.54 & 77.62 \\
     & & BLIP & \# images 129M &  \textcolor{lightgray}{78.25}  & \textcolor{lightgray}{78.32} \\
     & & SimVLM$_{base}$ & \# images 1.8B & \textcolor{lightgray}{77.87} & \textcolor{lightgray}{78.14} \\
    
    \cmidrule{2-6}

     & \multirow{9}{*}{SNLI-VE} & \textsc{Base Data-Size} &  & & \\
     & & UNITER  &  \# images 4M  & 79.39 & 79.38 \\
     & & UNIMO   &  \# images 4M  & \underline{81.11} & \underline{80.63} \\
     & & ALBEF   &  \# images 4M  & 80.14 & 80.30 \\
     & & REAVL (ours) & \# images 4M + Wikidata &  {\textbf {82.41}} & {\textbf {82.53}} \\
     \cmidrule{3-6}
     & & \textsc{Large Data-Size} &  \\
     & & ALBEF   & \# images 12M  & 80.80 & 80.91 \\
     & & SimVLM$_{base}$ & \# images 1.8B & \textcolor{lightgray}{84.20} & \textcolor{lightgray}{84.15} \\

    \bottomrule
    
\end{tabular}
\end{normalsize}
\end{center}
\vskip -0.1in
\end{table*}

\section{Experiment setup}

\subsection{Data} For the image-text data, following UNITER \cite{chen2020uniter}, we construct our pre-training data using two web datasets (Conceptual Captions, SBU Captions) and two in-domain datasets (COCO and Visual Genome). The total number of unique images is 4.0M, and the number of image-text pairs is 5.1M. We use Wikidata5M \cite{wang2021kepler}, a general-domain knowledge graph designed to capture background world knowledge for the KG data. It has 4M nodes and 20M edges in total. For each entity in Wikidata5M, we use its description collected from Wikipedia to produce entity embeddings.

\subsection{Implementation Details} Our model consists of a BERT-base with 123.7M parameters and a ViT-B/16 with 85.8M parameters. We pre-train the model for 10 epochs using a batch size of 32 on 8 NVIDIA A100 GPUs. 
For optimization, we use the AdamW \cite{loshchilov2017decoupled} optimizer with a learning rate of 5e-5 and a weight decay of 0.02.
We use DistMult \cite{yang2014embedding} for KG triplet scoring for the link prediction objective, with a negative exampling of 128 triplets and a margin of $\gamma = 0.05$. To pretrain the model, we perform MLM with a token masking rate of 25\%, MVM with a patch masking rate of 25\%, and link prediction with an edge drop rate of 15\%. Appendix \ref{supp:k_details} provides the details of the external knowledge memory and cost.

\subsection{Downstream evaluation tasks} We finetune and evaluate our models on the vision-language understanding tasks (OK-VQA, AOK-VQA, VQA-v2, and SNLI-VE) \cite{marino2019ok,schwenk2022okvqa,antol2015vqa,xie2019visual} and entity linking tasks (WikiDiverse and WikiPerson) \cite{wang2022wikidiverse,sun2022wikiperson}. For the vision-language understanding task, we follow the original setting and splits used by prior works \cite{li2022lavis}. Appendix \ref{downstream_task} provides the full details on these tasks and data.

\begin{table*}
    
    \centering
    \caption{Entity Linking Results on WikiDiverse and WikiPerson. R@K represents the recall of the TopK retrieved entities. The best and second-best results are marked {\bf number} and \underline{number}, respectively. }
    \vspace{0.05in}

    \begin{tabular}{c|ccc|ccc}
    \toprule
     &  
    \multicolumn{3}{c}{WikiDiverse} & 
    \multicolumn{3}{c}{WikiPerson} \\

     & R@10 & R@50 & R@100 &  R@1 & R@5 & R@10 \\
    \midrule
    ViT+BERT & 61.76 & 71.30 & 73.87 & 60.56 & 72.43 & 78.72 \\
    BLINK+CLIP\_N\_D & 66.96 & 77.18 & 80.53 & - & - & - \\

    ResNet+CLIP\_N  & 77.64 & 81.21 & 85.69 & 68.24 & 79.83 & 82.65 \\
    ResNet+CLIP\_N\_D & 80.64 & 84.33 & 87.56 & 73.70  & 83.47 & 84.45 \\
    
    CLIP & \underline{82.37} & \underline{87.82} & {\textbf{91.04}} & \underline{74.55} & \underline{84.42} & \underline{85.15} \\
    
    \midrule
    REAVL & {\textbf{83.20}} & {\textbf{88.42}} & \underline{89.59} & {\textbf{77.58}} & {\bf 85.27} & {\textbf{88.38}} \\
    
    \bottomrule
    
    \end{tabular}
    \label{tab:el_results}
\end{table*}

\begin{table}[t]
\caption{Ablation Study on OK-VQA.}
\label{ab-result}
\begin{center}
\begin{normalsize}
\begin{sc}
\begin{tabular}{lcc}
\toprule
Training Task / Architecture & OK-VQA\\
\midrule
 \textup{w/o Knowledge Retriever}     &   52.71  \\
 \textup{w/o GNN Aggregation}  &   54.33  \\
 \textup{w/o Knowledge Augmented Model}   &  56.61  \\

\midrule
 \textup{Retrieval From Textual}  &   53.27    \\
 \textup{Retrieval From Vision (Final)}   &   {\textbf{57.72}}   \\

\midrule
 MLM        &    54.38  \\
 MLM+MVM     &   55.60   \\
 MLM+MVM+ITC &   57.24   \\
 MLM+MVM+ITC+LinkPred    &   {\textbf{57.72}}   \\

\bottomrule
\end{tabular}
\end{sc}
\end{normalsize}
\end{center}
\end{table}

\section{Experiment Results}

\subsection{Evaluation on V+L understanding tasks}

To examine the quality of multi-modal representation learning, we first compare REAVL on four downstream V+L tasks with state-of-the-art methods, as shown in Table \ref{res-table}. 
It is clear that our method REAVL achieves state-of-the-art performance, outperforming all existing models including supervised models (KAT, REVIVE) and VLP models (ALBEF, BLIP) on the knowledge-based task. For example, on the OK-VQA dataset, REAVL achieves a 1.94\% and 4.15\% relative accuracy gain over the baseline REVIVE and BLIP respectively. These improvements are consistent in the AOK-VQA dataset, demonstrating our hypothesis that large knowledge graphs can provide complementary information to multi-modal data. 

Particularly, compared to the strong baseline REVEAL, which also exploits much knowledge to improve multi-modal representation, our model shows superior performance under the same size of parameters with fewer knowledge resources, demonstrating the superiority of our self-supervised learning scheme for the retrieval of informative knowledge. This is in line with our expectations as our self-supervised learning scheme could efficiently identify informative knowledge from large-scale knowledge memory and integrate them with visual and textual information.

For the general V+L tasks, our model REAVL achieves relative improvements of 3.34\% on VQA-v2 and 2.36\% on SNLI-VE, with 4M pre-training images. When comparing with the models that are trained with a significantly larger number of data, our model also shows competitive performances. For example, our model outperforms ALBEF and BLIP pre-trained on $>$10M data on both datasets. The performance supports our view that the results of the SOTA model under larger pre-training data can be achieved through the fusion of smaller data scales and knowledge. And the slightly lower performance compared to SimVLM is expected since larger data (1.8B) can provide more information than knowledge graphs. 

Similar results have been shown on other general vision-language tasks such as Image Captioning, as shown in Appendix \ref{ref:ic_task}. With a small amount of data, our model achieves improvements over BLIP of 4.39\% on COCO Captions and 1.62\% on NoCaps. When comparing with the models that are trained with a significantly larger number of data, our model also shows competitive performances.

\begin{table*}[t]
\caption{Case Study on OK-VQA dataset.}
\label{case-table}
\begin{center}
\begin{normalsize}
\begin{tabular}{cccccc}
\toprule
 \textsc{Question} & \textsc{Image} & \textsc{Ground-Truth} & \textsc{Prediction} & \textsc{Supporting} \\
\midrule

\makecell[c]{(a) What is the person \\in the photo wearing?} &\begin{minipage}[htbp]{0.3\columnwidth}
		\centering
		\raisebox{.05\height}{\includegraphics[width=\linewidth]{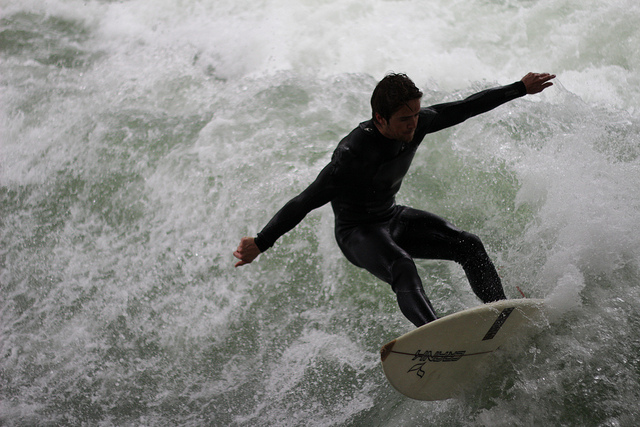}}
	\end{minipage}
  &  \makecell[c]{[wetsuit, suit,\\wet suit, trunk]}
 & \textsc{REAVL}: wetsuit \textcolor{ForestGreen}{\ding{51}} & 
 \makecell[l]{
     \textbf{Retrieved Entities}:\\  \{boardsport, surfer, big wave \\surfing, \textcolor{ForestGreen}{\bf wetsuit}, dry suit, ...\} \\
 }
 \\

 \makecell[c]{(b) What kind of flowers \\are on the table?
}  &\begin{minipage}[htbp]{0.3\columnwidth}
		\centering
		\raisebox{0.05\height}{\includegraphics[width=\linewidth]{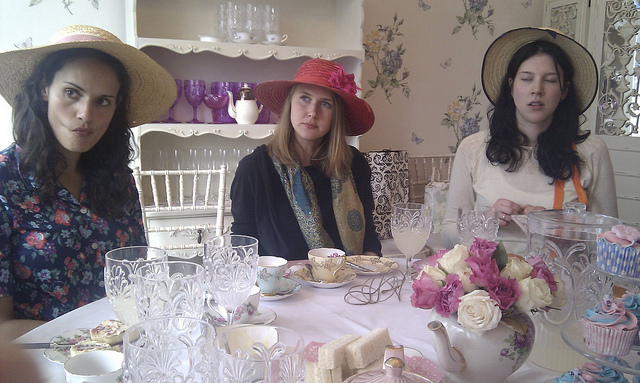}}
	\end{minipage}
  &  \makecell[c]{[rose]} & \textsc{REAVL}: rose \textcolor{ForestGreen}{\ding{51}} & 
 \makecell[l]{
     \textbf{Retrieved Entities}:\\  \{dinnerware, Duchess of \\Abrantes, cake plate, tea,\\
     \textcolor{ForestGreen}{\bf rose petals}\}
 }
\\

\makecell[c]{(c) Which liquid seen here \\comes from a citrus fruit? } & \begin{minipage}[htbp]{0.3\columnwidth}
		\centering
		\raisebox{0.05\height}{\includegraphics[width=\linewidth]{case/COCO_val2014_000000331817.jpg}}
	\end{minipage}
  & \makecell[c]{[orange juice, orange]} &  \textsc{REAVL}: juice \textcolor{Red}{\ding{53}} &  
 \makecell[l]{
     \textbf{Retrieved Entities}:\\  \{plateau, placemat, \textcolor{ForestGreen}{\bf fruit}\\\textcolor{ForestGreen}{\bf juice}, monkey bread, \textcolor{ForestGreen}{\bf citrus} \\\textcolor{ForestGreen}{\bf juice}, ...\}
 }
 \\

  \makecell[c]{(d) Can you guess the location \\where the airoplane is seen?
}  &\begin{minipage}[htbp]{0.3\columnwidth}
		\centering
		\raisebox{0.05\height}{\includegraphics[width=\linewidth]{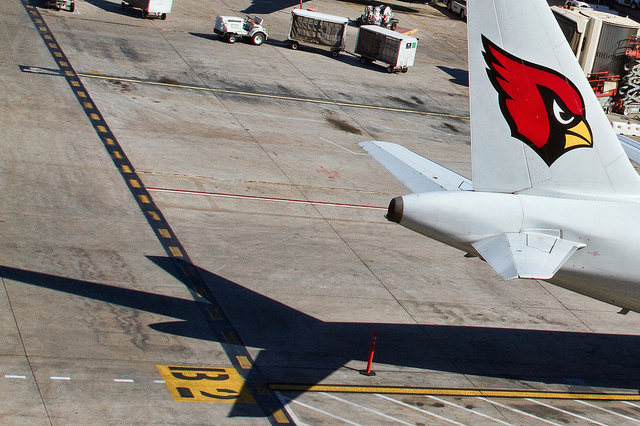}}
	\end{minipage}
  & \makecell[c]{[arizona, new zea-\\land, us, tarmac]} & \textsc{REAVL}: airport \textcolor{Red}{\ding{53}} &  
 \makecell[l]{
     \textbf{Retrieved Entities}:\\  \{Canadair Regional Jet,\\forward-swept wing, yellow\\line, non-towered \textcolor{ForestGreen}{\bf airport}, ...\}
 }
 \\

\midrule
\makecell[c]{(e) What type of\\ temperature is this?} & \begin{minipage}[htbp]{0.3\columnwidth}
		\centering
		\raisebox{0.05\height}{\includegraphics[width=\linewidth]{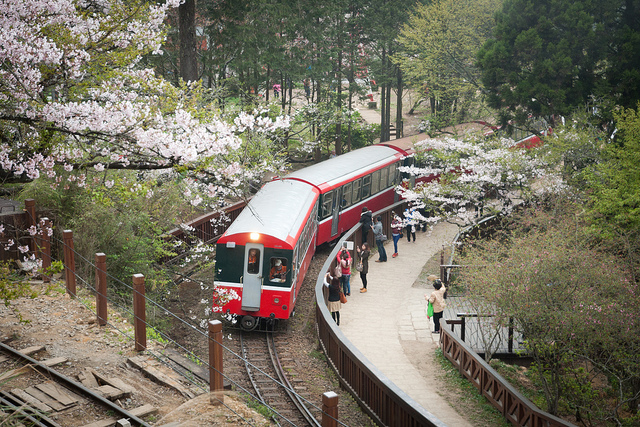}}
	\end{minipage}
  & \makecell[c]{[cold, cold, cold,\\cold, cold, cold,\\ cold, cold, pleasant,\\pleasant]} & \makecell[c]{
      \textsc{REAVL}: cold \textcolor{ForestGreen}{\ding{51}} \\
      \textsc{BLIP}: cool \textcolor{Red}{\ding{53}} \\ 
      \textsc{REVIVE}: warm \textcolor{Red}{\ding{53}} 
  } &  
 \makecell[l]{
     \textbf{Retrieved Entities}:\\  \{Strawberry train, Nankai\\Electric Railway (Japanese\\railway company), rail\\transport in Walt Disney Parks,\\\textcolor{ForestGreen}{\bf cherry blossom}\}
 }
 \\

\bottomrule
\end{tabular}
\end{normalsize}
\end{center}
\end{table*}

\subsection{Evaluation on Entity Linking tasks}

REAVL is also capable of linking the visual mentions in the image to the corresponding named entity in knowledge graphs. To demonstrate the ability, we compared REAVL with the existing multi-modal entity linking models.
We followed the evaluation settings used by \cite{wang2022wikidiverse} and \cite{sun2022wikiperson} to report the comparison results with existing methods in Table \ref{tab:el_results}.
Considering the format of entities in KBs, we only consider the visual (image) to textual (descriptions and names) entity linking.

As we can see, ViT+BERT has achieved reasonably good performance for R@10 (i.e., 61.76 at WikiDiverse and 78.72 at WikiPerson), which demonstrates the effectiveness of the pre-training model. Moreover, we can see that the CLIP, which is pre-trained with about 400M image-caption pairs, has achieved the strongest baseline performances, which demonstrates the effectiveness in combining both visual description and textual description. Our model has outperformed the CLIP model on 5 out of 6 metrics across the two datasets, proving that our pre-training objectives on multi-modal data such as MLM and MVM have promoted the training of the knowledge retriever. The slightly lower performance in R@100 is expected as we select candidate entities within the Top-50 for training due to the limitation of the length of the knowledge-augmented model.

\subsection{Ablation Study}
Here, we will ablate the properties of REAVL to investigate factors that influence the model performance, as we discuss below. Table \ref{ab-result} and Appendix \ref{supp:abla} show the ablation experiment results.

\subsubsection{\textbf{Incorporating Knowledge Graphs with Multi-modal data.}}
The first key contribution of our model (w.r.t. existing VLP methods) is that we incorporate knowledge graphs for V+L pre-training. 
We find that this significantly improves the model’s performance for knowledge retrieval and integration.  
Compared to the REAVL without the KG retriever, jointly training the KG retriever and V+L learning substantially improves the model’s performance on the OK-VQA task. Furthermore, adding the proposed GNN Aggregation and the knowledge-augmented model both enhance the model performance. This is in line with our expectation as GNN Aggregation could extract more related entities and relations while the knowledge-augmented model is designed to fuse knowledge and multi-modal data in an efficient way.

\subsubsection{\textbf{Knowledge retrieval on multi-modal data}}

Another advantage of our model is its ability to retrieve knowledge from multiple modalities like images. In order to study the necessity of visual knowledge retrieval, we perform an ablation study to see the impact of knowledge retrieval. As can be seen, the performance drop of textual retrieval is more severe on the OK-VQA dataset. This is understandable because images contain more information that is not covered by text, especially in the downstream tasks. Thus, it is necessary to retrieve knowledge from images to better facilitate the fusion of knowledge and multi-modal data.

\subsubsection{\textbf{Effect of pre-training objectives}}

In order to study the contributions of different pre-training objectives, we investigated the four pre-training objective combinations. We
report their fine-tuned results on the OK-VQA dataset in Table \ref{ab-result}. As can be seen, only with MLM, our model only achieves an accuracy of 54.38, which slightly lags behind the baseline models (ALBEF and BLIP). Adding MVM and ITC both substantially improves the pre-trained model’s performance. And the proposed LinkPred further enhances the model by reasoning complex world knowledge. Finally, our model achieves a 1.94\% and 4.15\% 
relative accuracy gain over the baseline REVIVE and BLIP respectively. The ablation study demonstrates the effectiveness and complementarity of each self-supervised task.

\subsection{Case Study}

We conduct case studies on the behavior of REAVL’s knowledge retrieval and knowledge augmentation, where we analyze the contribution of retrieved entities to question answering (Table \ref{case-table}). 
We can observe that our approach can accurately retrieve informative entities after training the knowledge retriever. For instance, REAVL can retrieve useful knowledge from images (e.g., wetsuit and rose petals) to generate the correct answer in example (a)(b). From the error cases, we can see that the model still generates reasonable answers for such scenarios. For example (c), we have retrieved the informative entity \textit{citrus juice} from the image, but failed to understand the problem well. Example (d) is quite difficult to answer but our model still could generate a reasonable answer \textit{airport}. When comparing the existing methods, we selected the case that has been reported by REVIVE. Although the question is more complex, our model could accurately predict the answer, but existing models, BLIP and REVIVE, struggle to predict the correct answers, which can demonstrate the potential of our proposed method.

\section{Conclusion}
In this paper, we presented REAVL, a novel vision-language pre-training framework to incorporate world knowledge into multi-modal representations. It not only exploits the multi-modal data for better knowledge retrieval but also fuses knowledge and multi-modal data with a knowledge-augmented model. Our experiments on knowledge-based tasks show the superiority of our approach, as it outperforms existing VLP baselines with low-resource pre-training data. 
At the same time, the performance on entity linking tasks also shows its excellent ability in retrieving informative knowledge from massive knowledge graphs. 
One limitation of REAVL is that it also constitutes large computational costs due to knowledge retrieval, and an important future research will be to extend it to make it faster and apply it to larger datasets.

\begin{acks}
This work is supported by the National Key R\&D Program of China (2020YFB0204803); National Natural Science Foundation of China (12126610); Guangdong Key Field R\&D Plan (2019B020228001 and 2018B010109006); Guangzhou S\&T Research Plan (202007030010 and 202002020047).
\end{acks}

\clearpage
\bibliographystyle{ACM-Reference-Format}
\balance
\bibliography{sample-base}


\clearpage
\appendix
\section{Knowledge details \label{supp:k_details}}

\subsection{Knowledge Retriever. \label{supp:k_retriever}}
We initialized our visual encoder $f_v$ by CLIP. And for the retriever, we just build the MIPS index once for simplicity and do not update the entity embedding $f_e$ at the retriever. Note that we still fine-tune the visual embedding $f_v$, so we do not perform the asynchronous index refresh as in REALM fine-tuning. It is possible that refreshing the index would further improve performance. And we update our entity embedding $f_e$ at the GNN and knowledge-augmented model.



\begin{table*}[t]
\caption{Knowledge Extra Cost}
\label{cost-table}
\begin{center}
\begin{small}
\begin{sc}
\begin{tabular}{ccccc}
    \toprule
    & \# instance 	& \# triplets	& storage (initial-embedding)	& storage (faiss-index) \\
    Wikidata5M	& 4,594,485	& 20,614,279	& 8.76GB	& 2.19GB \\
    \bottomrule
\end{tabular}
\end{sc}
\end{small}
\end{center}
\end{table*}


\subsection{Knowledge Extra Cost \label{supp:k_cost}}
The extra cost of the knowledge-augmented REAVL model is most from the knowledge retriever. And we use an efficient retrieval method thanks to the FAISS. During the pre-training, we took 8 days for pre-training on 8 NVIDIA A100 GPUs with a batch size of 32 and an epoch of 20, which is slightly slower than baseline methods (ALBEF and BLIP) using the same machine. As for inference, our model took 2s per sample for VQA answer generation, compared to 1.8s of the baseline method BLIP on the same machine.



\section{Data \label{downstream_task}}

\subsection{V+L understanding tasks} requires a cooperative knowledge of vision and language. We finetune and evaluate four vision-language understanding benchmarks, including knowledge-base tasks (OK-VQA \cite{marino2019ok} and AOK-VQA \cite{schwenk2022okvqa}) and general tasks (VQAv2 \cite{antol2015vqa} and SNLI-VE \cite{xie2019visual}). Specifically, we feed the embeddings of a given question (text), an image, and the retrieved entities into the knowledge-augmented model and use a 12-layer transformer decoder to generate the answer. 

For the VQA-v2, OK-VQA, and AOK-VQA, we followed the prior work \cite{li2021align,li2022lavis} and formulate the task as a classification problem over the most frequent answers in the training set. OK-VQA and AOK-VQA is the knowledge-based VQA benchmark that requires external knowledge to answer its questions, that is, the knowledge that is not directly present in the image input.



\subsection{Multi-modal Entity Linking} aims at linking mentions with multi-modal contexts to the referent entities from a knowledge base (e.g., Wikipedia), is an essential task for many multi-modal applications. We consider two datasets for multi-modal Entity Linking: WikiDiverse \cite{wang2022wikidiverse} and WikiPerson \cite{sun2022wikiperson}. WikiDiverse is a high-quality human-annotated
MEL dataset with diversified contextual topics and entity types from Wikinews, while WikiPerson is a high-quality human-annotated visual person-linking dataset from VisualNews. Herein, we finetuned the dual encoder of image and entity with the contrastive loss function of mention-entity samples to demonstrate the superiority of our retrieval process.

\section{Additional Experiments}

\begin{table*}[t]
\caption{Results for vision-language pre-training methods on popular image captioning benchmarks. We report CIDEr for COCO Captions and NoCaps. The best and second-best results are marked {\bf number} and \underline{number}, respectively. The gray \textcolor{lightgray}{number} indicates that the model is trained with a significantly larger number of data than our models. }
\label{res-ic-table}
\begin{center}
\begin{normalsize}
\begin{sc}
\begin{tabular}{lc|cc}
    \toprule
    Model & Knowledge Resources & COCO Caption & NoCaps \\
    \midrule
     Base Data-Size &  \\
     BLIP & \# image 14M &	129.7	& 105.1 \\
     REVAL &	\# image 4M + Wikidata5M	&   \textbf{135.4}	& \textbf{106.8} \\
     
     \midrule
     Large Data-Size &  \\
     SimVLM-base &	\# image 1.8B	&134.8	&94.8 \\
     SimVLM-large &	\# image 1.8B	& \textcolor{lightgray}{142.6}	& \textcolor{lightgray}{108.5} \\
     SimVLM-huge & \# image 1.8B	    & \textcolor{lightgray}{143.3}  & \textcolor{lightgray}{110.3} \\

    \bottomrule
\end{tabular}
\end{sc}
\end{normalsize}
\end{center}
\end{table*} 






\begin{table*}[t]
\caption{Additional Results for the entity linking task. (a) Compared with the text-based methods on WikiDiverse; (b) Zero-shot learning result on WikiPerson.}
\label{tab:zero-shot}
\vspace{-5mm}
\begin{center}
\begin{small}
\begin{sc}
\begin{tabular}{lccc}

    \toprule

     & Recall@10 & Recall@50 & Recall@100  \\
    \midrule
    BLINK & 63.63 & 73.15 & 76.03 \\
    REAVL & {\textbf{83.20}} & {\textbf{88.42}} & \textbf{89.59}  \\

    \bottomrule

    & Zero-Shot R@1	& Zero-Shot R@5	& Zero-Shot R@10 \\
    \midrule
    CLIP	& 54.41	& 68.42	& 75.32 \\
    REVAL	& \textbf{55.13}	& \textbf{69.24}	& \textbf{76.39} \\
    \bottomrule
\end{tabular}
\end{sc}
\end{small}
\end{center}
\end{table*} 

\begin{table*}[htp]
\caption{Case study of Knowledge retriever.}
\label{supp-table:retriever}
\begin{center}
\begin{normalsize}
    
\begin{tabular}{lccc}
    \toprule
    Model & \textsc{Question} & \textsc{Image} & Retrieved Knowledge \\
    \midrule
     
     CLIP & \makecell[c]{What is the person \\in the photo wearing?} &	\begin{minipage}[htbp]{0.3\columnwidth}
		\centering
		\raisebox{.05\height}{\includegraphics[width=\linewidth]{case/COCO_val2014_000000319073.jpg}}
	\end{minipage}	&  
 \makecell[l]{
     \{"wave energy",  "wakeboarding resort", \\ "Dibyashwori Hydropower Ltd. ", \\ "Surfer Riding a Wave" \\ "waterist", \}
 }  \\
 
     REVAL &	\makecell[c]{What is the person \\in the photo wearing?}	&   \begin{minipage}[htbp]{0.3\columnwidth}
		\centering
		\raisebox{.05\height}{\includegraphics[width=\linewidth]{case/COCO_val2014_000000319073.jpg}}
	\end{minipage}	& 
 \makecell[l]{
     \{"boardsport", "surfer", "big wave \\ surfing", "wetsuit", "dry suit"\}
 } 
  \\
     
    \bottomrule
\end{tabular}
\end{normalsize}
\end{center}
\end{table*}

\subsection{Experiments on Image Captioning Task. \label{ref:ic_task}} 
We also evaluated our model on other general vision-language tasks such as Image Captioning. The results are consistent with the VQA and VE datasets in trend, as shown in Table \ref{res-ic-table}. With a small amount of data, our model REAVL achieves improvements over BLIP of 4.39\% on COCO Captions and 1.62\% on NoCaps. When comparing with the models that are trained with a significantly larger number of data, our model also shows competitive performances.

\subsection{Zero-shot Learning on entity linking task. \label{supp:zero_shot}}
We add the experiments of the zero-shot tasks on the WikiPerson dataset to demonstrate the ability of the retriever, as shown in Table \ref{tab:zero-shot}. The results demonstrate the zero-shot ability of our model and the improvement over CLIP also illustrates that our knowledge retriever has been learning in the correct direction.

\subsection{Ablation study}
\label{supp:abla}

\subsubsection{Ablation study on the number of retrieved knowledge entries. \label{supp:topk-k}}
We have conducted the ablation study on the impact of the number of retrieved knowledge entries (Top-K). As the number of retrieved knowledge entries increases, the performance of REAVL increases rapidly thanks to the more informative knowledge. Subsequently, the performance slowly descends as K continues to increase, indicating that an overlarge of retrieved knowledge entries will bring redundant and harmful information.


\subsubsection{Analysis on Knowledge Retriever. \label{supp:retriever}}
As shown in Table \ref{supp-table:retriever}, we analyzed the prediction results on OK-VQA to determine whether they were caused by the error of the retriever. For instance, we compared the CLIP model with our knowledge retriever when answering the question "What is the person in the photo wearing?". CLIP model could recognize multiple objects in the image, such as waterist, wave energy, and even a company in Nepal. But the retrieved knowledge is incorrect as it does not relate to the question text, resulting in the wrong answer. In contrast, our knowledge retriever, with the help of the proposed knowledge-aware task, makes an accurate prediction by focusing on the person in the image wearing a "wetsuit" or "dry suit". It again verified the importance of our proposed knowledge-aware self-supervised tasks.

For the GNN Aggregation and Knowledge Augmented module, as shown in example (e) in Table 4, to answer the question "What type of temperature is this?", the knowledge from the input image and question is not sufficient for answering. And our methods could retrieve the entity "cherry blossom" from an image and predict the correct answer "warm" based on the climate in which cherry blossoms usually bloom. Interestingly, the useful knowledge does not directly come from the retrieved entity but rather from its neighboring knowledge ("Blooming season"). In contrast, the baseline method REVIVE without the GNN Aggregation module cannot predict the correct answer. It again verified the importance of our proposed GNN Aggregation and Knowledge Augmented module.


\subsubsection{Model Scalability. \label{supp:m_scale}}

We have demonstrated that our model can be scaled to more data. Table \ref{supp-table:scalability} shows the performance comparison of different data sizes. The significant improvement demonstrates the scalability of our model. We also compared the ALBEF model with our KG modeling using the same 1.3M (in-domain, COCO+VG) dataset. Both models were trained with a batch size of 32 using the same number of machines. The ~8\% improvements presented in the table below demonstrate the superiority of our KG modeling in the same data regime.


\begin{minipage}[c]{0.5\textwidth}
\makeatletter\def\@captype{table}
\centering
\caption{Model Scalability}
\label{supp-table:scalability}
\begin{center}
\begin{normalsize}
\begin{sc}
\begin{tabular}{lcc}
    \toprule
    
    Model & Pretraining-Data-size &   OK-VQA \\
    \midrule
    ALBEF &	CC(in domain 1.3M)  &   40.7 \\
    REAVL & CC(in domain 1.3M) &	48.6 \\
    REAVL & CC4M &	\textbf{53.4}  \\
    \bottomrule
    
\end{tabular}
\end{sc}
\end{normalsize}
\end{center}
\end{minipage}

\begin{minipage}[c]{0.4\textwidth}
\makeatletter\def\@captype{table}
\centering
\caption{Ablation study on the number of retrieved knowledge entries (Top-K).}
\label{suppl-table:topk}
\begin{center}
\begin{normalsize}
    
\begin{sc}
\begin{tabular}{lc}
    \toprule
    Model &	WikiDiverse \\
    & R@10 \\
    \midrule
    REAVL (K=10)	& 78.39 \\
    REAVL (K=20)	& 80.57 \\
    REAVL (K=50)	& \textbf{83.20} \\
    REAVL (K=100)	& 82.96 \\
    \bottomrule
\end{tabular}
\end{sc}
\end{normalsize}
\end{center}
\end{minipage}










\end{document}